\documentclass{article}

\usepackage[nonatbib,preprint]{nips_2018}

\usepackage[utf8]{inputenc} 
\usepackage[T1]{fontenc}    
\usepackage{hyperref}       
\usepackage{url}            
\usepackage{booktabs}       
\usepackage{amsfonts}       
\usepackage{nicefrac}       
\usepackage{microtype}      
\usepackage{graphicx}
\usepackage[ruled,vlined,linesnumbered]{algorithm2e}
\usepackage{subcaption}
\usepackage{multirow}
\usepackage{adjustbox}

\usepackage[
  separate-uncertainty = true,
  multi-part-units = repeat
]{siunitx}

\title{Detecting Adversarial Samples for Deep Neural Networks through Mutation Testing}

\author{
  Jingyi~Wang, Jun Sun\\
  Information Systems Technology and Design\\
  Singapore University Of Technology and Design\\
  \And
  Peixin Zhang, Xinyu Wang \\
  College of Computer Science and Technology\\
  Zhejiang University \\
}

\begin{document}

\maketitle

\begin{abstract}
Recently, it has been shown that deep neural networks (DNN) are subject to attacks through adversarial samples. Adversarial samples are often crafted through adversarial perturbation, i.e., manipulating the original sample with minor modifications so that the DNN model labels the sample incorrectly. Given that it is almost impossible to train perfect DNN, adversarial samples are shown to be easy to generate. As DNN are increasingly used in safety-critical systems like autonomous cars, it is crucial to develop techniques for defending such attacks. Existing defense mechanisms which aim to make adversarial perturbation challenging have been shown to be ineffective. In this work, we propose an alternative approach. We first observe that adversarial samples are much more sensitive to perturbations than normal samples. That is, if we impose random perturbations on a normal and an adversarial sample respectively, there is a significant difference between the ratio of label change due to the perturbations. Observing this, we design a statistical adversary detection algorithm called \emph{nMutant} (inspired by mutation testing from software engineering community). Our experiments show that \emph{nMutant} effectively detects most of the adversarial samples generated by recently proposed attacking methods. Furthermore, we provide an error bound with certain statistical significance along with the detection. 
\end{abstract}

\section{Introduction}
\begin{figure}[t]
    \centering
    \begin{subfigure}[b]{0.45\textwidth}
        \includegraphics[width=\textwidth]{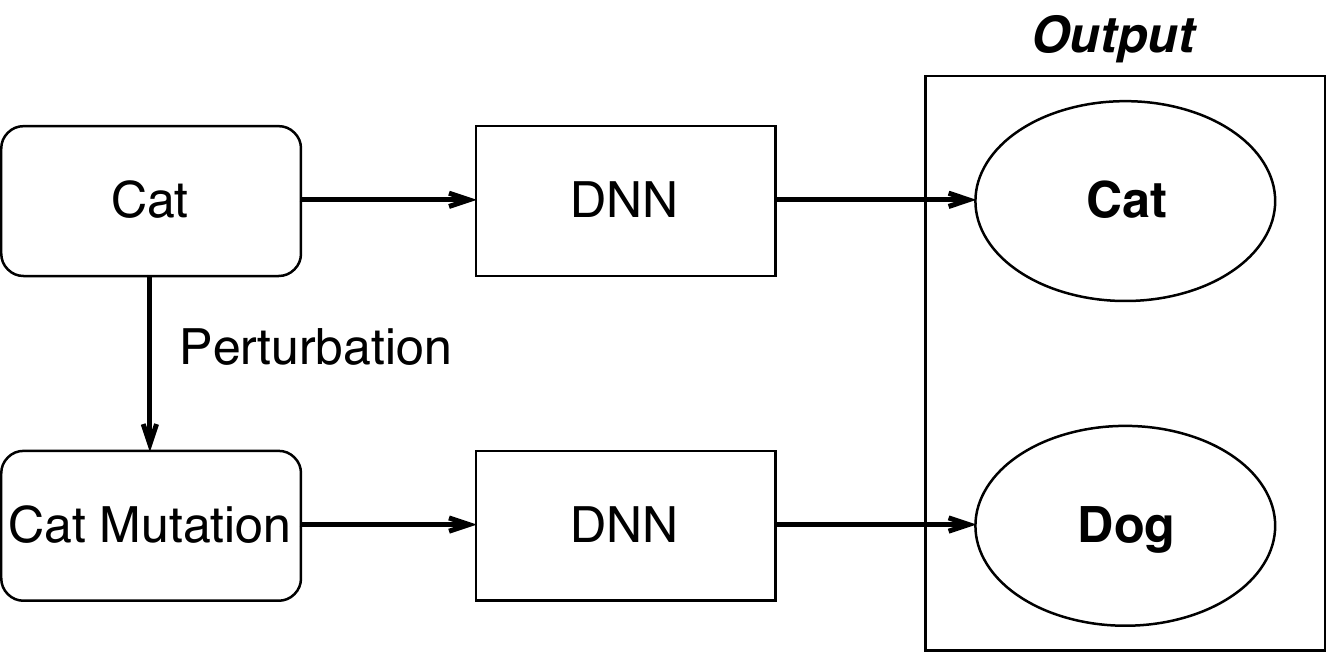}
        \label{fig:class:normal}
    \end{subfigure}
    \hfill 
    \begin{subfigure}[b]{0.45\textwidth}
        \includegraphics[width=\textwidth]{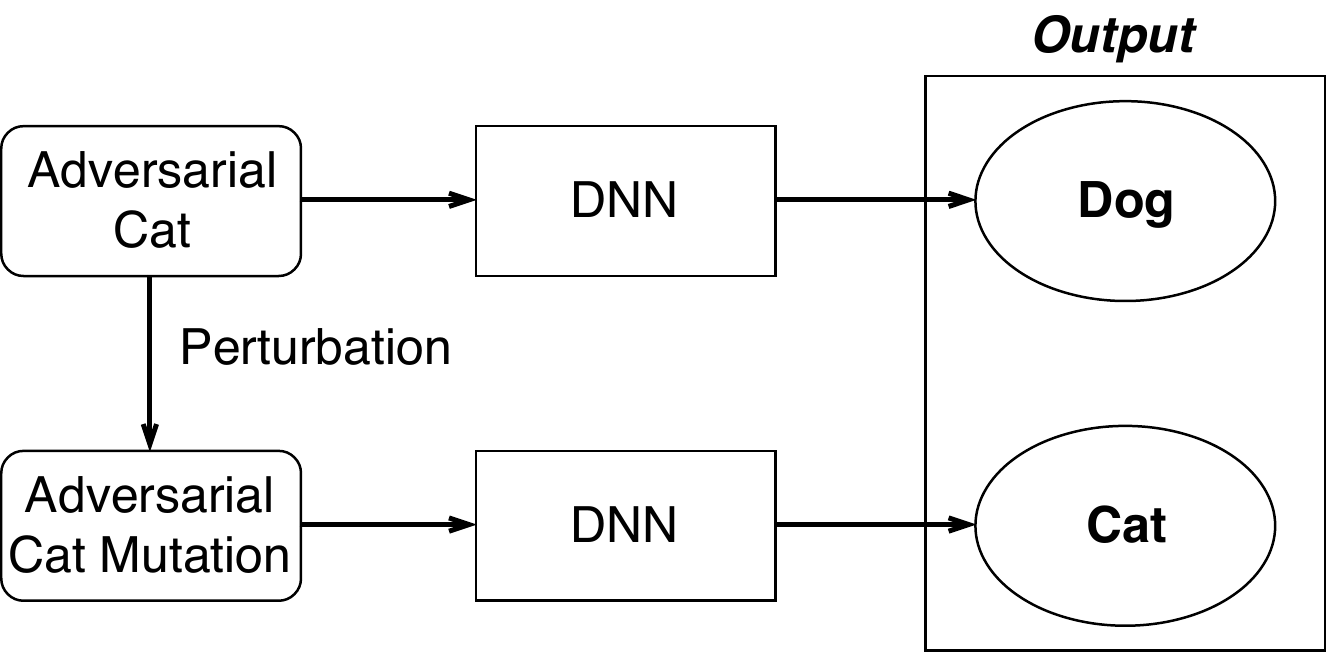}
        \label{fig:class:adv}
    \end{subfigure}
  \caption{Label change via perturbation on a normal sample (left) and an adversarial sample (right).}
  \label{fig:class}
\end{figure}

Deep Neural Networks (DNN) have been applied in a wide range of applications in recent years and shown to extremely successful in solving problems. However, it has also been shown that even well-trained DNN can be vulnerable to attacks through adversarial samples, especially when DNN are applied to classification tasks~\cite{goodfellow2014explaining,moosavi2016deepfool}. Adversarial samples are created with the intent to trigger errors of the DNN. They are often crafted through adversarial perturbation, i.e., manipulating the original sample with minor modifications so that the DNN model labels the sample incorrectly. This is illustrated by the left part of Fig.~\ref{fig:class}.
Many methods have been invented recently to craft such adversarial samples, e.g., fast gradient sign method~\cite{goodfellow2014explaining} and its variants~\cite{moosavi2016deepfool}, Jacobian-based saliency map approach~\cite{papernot2016limitations}, C\&W L2 attack~\cite{chen2017ead} and so on~\cite{su2018attacking,sharif2016accessorize,galloway2017attacking,brendel2017decision,xiao2018spatially,papernot2017practical}.

As DNN are increasingly used in safety-critical systems like autonomous cars, it is crucial to develop effective techniques for defending such attacks.
To counter attacks through adversarial samples, multiple defense strategies have been proposed with the aim to improve the robustness of DNN. Existing defenses are largely either based on the idea of adversary training~\cite{tramer2017ensemble,papernot2016distillation,papernot2017extending,gu2014towards,cisse2017parseval,guo2017countering,athalye2018obfuscated,shaham2015understanding,sinha2018certifying} which works by taking adversarial samples into consideration during model training (so that it is harder for the attackers to craft adversarial samples), or based on detecting adversary samples by training a subsidiary model~\cite{metzen2017detecting,xu2017feature,grosse2017statistical}. We contend that most existing defense strategies are dependent on the available adversarial samples, and thus are usually limited to defend specific attacks. That is, they provide no guarantee or confidence if the DNN is faced with a new attack. Furthermore, it is shown in~\cite{huang2017safety,katz2017reluplex,katz2017towards} that formally verifying DNN to provide safety guarantees is too computationally expensive and thus would not scale.


In this work, we propose an alternative approach for efficiently detecting adversarial samples at runtime. \emph{Our approach is based on the observation that adversarial samples are much more sensitive to random perturbations than normal samples (i.e., those which are correctly labeled by the DNN).}
That is, the probability of obtaining a different label by imposing random perturbations to an adversarial sample is significantly higher that of imposing random perturbations to a normal sample. This is illustrated in Fig.~\ref{fig:class}, i.e., the probability of the scenario illustrated on the right happening is significantly more than that of the scenario illustrated on the left happening. We confirm this observation through an empirical study with standard datasets and recently proposed adversarial perturbation methods (refer to Section~\ref{exp} for details). This observation can be intuitively explained through a simple explanatory model.

Based on the observation, we propose the method for adversarial sample detection, which is inspired by mutation testing developed in software engineering community. The basic idea is to measure how sensitive a provided sample is to random perturbations and raise an alarm if the sensitivity is above certain threshold. Through empirical study on the MNIST 
and CIFAR10 dataset, we show that our approach is effective against many existing attacking methods (e.g., FGSM~\cite{goodfellow2014explaining}, C\&W~\cite{carlini2017towards}, JSMA~\cite{papernot2016limitations}, and Blockbox~\cite{papernot2017practical}). Our approach works without requiring any knowledge of the underlying DNN system and thus can be potentially applied to a wide range of systems. It is reasonably scalable
and reliable (by providing a confidence along with the detection) compared to existing defense strategies. We frame the rest of the paper as follows. We discuss related works in Section~\ref{rel}. Then we formally define our adversarial sample detection problem in Section~\ref{probdef}. Our observation and proposed detection algorithm are presented in Section~\ref{mt} and Section~\ref{defense} respectively. Lastly, we present experiment results in Section~\ref{exp}.

\section{Related works}\label{rel}

Due to the fast growing interests in this research area, many works are emerging which we are not able to completely cover. Thus, we focus on the most relevant works in the following.

First of all, our work is devised to counter the many recently proposed attacks on DNN. These attacks can be roughly divided into two categories, i.e., white-box attacks with access to the DNN~\cite{goodfellow2014explaining,moosavi2016deepfool,pei2017deepxplore,xiao2018spatially,carlini2017towards} and black-box attacks without the knowledge of the DNN~\cite{papernot2016transferability,papernot2017practical,liu2016delving,mopuri2017fast}. Our adversarial sample detection algorithm does not require the knowledge of the DNN and thus works in a black-box fashion. It can be applied to defend both white-box attacks and black-box attacks.


On the defending side, one main line of work to improve the robustness of DNN is by adversarial training, which augments training data with adversarial data and modifies the training phase so that it is harder for the attackers to craft adversarial samples~\cite{szegedy2013intriguing,goodfellow2014explaining,kurakin2016adversarial,madry2017towards,tramer2017ensemble,papernot2016distillation,papernot2017extending,gu2014towards,cisse2017parseval,guo2017countering,athalye2018obfuscated,shaham2015understanding,sinha2018certifying}. Besides adversarial training, another line of defense is to detect adversarial samples by training subsidiary models from adversarial data~\cite{metzen2017detecting,xu2017feature} or testing the statistical difference between the training data and adversarial data~\cite{grosse2017statistical}.

There are some relevant research in the software engineering community as well. Both white-box and black-box testing strategies have been proposed to generate adversarial data more efficiently~\cite{pei2017deepxplore,wicker2018feature,tian2017deeptest}. There are also attempts to formally verify the DNN to provide safety and security guarantees~\cite{huang2017safety,katz2017reluplex,katz2017towards}, which are however proven to be too computationally expensive and not scalable to large DNN.

Our work falls into the category of adversary detection. However, compared to adversarial training and prior adversary detection algorithms, our algorithm does not rely on the available adversarial data. Furthermore, we are able to report a confidence along with the detection. Compared to formal verification of DNN, our approach is scalable because it is based on statistical testing and randomly mutating the given sample.

\section{Problem definition}\label{probdef}

We denote the target DNN by $f(X):X\to C$, where $X$ is the set of input samples and $C$ is the set of output labels. In this work, we assume that we do not have any knowledge of $f$ other than that we can obtain its output (i.e., the label) given a certain sample. Given a sample $x$, we denote its true label (obtained by human observer) by $c_x$. We say that a sample $x$ is a normal sample if $f(x)=c_x$ and is an adversarial sample if $f(x) \neq c_x$. Notice that according to our definition, a sample in the training data which is wrongly-labeled is also an adversarial sample. Our problem is then, given an arbitrary sample $x$ and the DNN $f$, how can we effectively detect whether $x$ is a normal sample or an adversarial sample? Can we provide some confidence for the detection result as well?

\section{Mutation testing}\label{mt}

We first present our observation on the sensitivity of mutation testing on a normal sample and an adversarial sample. Given an arbitrary sample $x$ for a DNN, 
we obtain a set of mutations $X_m(x)$, each of which $x_m=x+\epsilon$ is obtained by imposing a minor \textit{random} perturbation $\epsilon$ on $x$ that is label-preserving. We remark that we restrict the perturbations in domain specific ways to guarantee that the mutations are realistic~\cite{pei2017deepxplore}. For example, for image recognition systems, the perturbation can be lighting effect to simulate different intensity of lights and occlusion of a rectangle to simulate a blocking area of a camera, etc. For every mutation $x_m\in X_m(x)$, we obtain its output label $f(x_m)$ by feeding $x_m$ into the DNN.

Intuitively, for most of the mutations in $X_m(x)$, $f(x_m)$ should be $f(x)$ because we are imposing a small, realistic and thus hopefully label-preserving perturbation on $x$. As demonstrated in previous works~\cite{szegedy2013intriguing,goodfellow2014explaining,kurakin2016adversarial}, however, a label change might occur. An example of label change is illustrated in Figure~\ref{fig:class}. The left figure takes a normal `cat' image as input and the DNN correctly classifies it as a `cat'. After imposing a random perturbation, the DNN labels it as a `dog'. The right figure however takes an adversarial `cat' image (which could be either due to  adversarial perturbation or training error) as input and the DNN wrongly labels it as a `dog'. Then, after imposing a random perturbation, the DNN labels it as a `cat'.

We calculate the sensitivity of a sample $x$ to perturbations as follows:
\[\kappa(x) = \frac{|\{x_m|x_m\in X_m(x) \land f(x_m)\neq f(x)\}|}{|X_m(x)|}\]
, where $|S|$ is the number of elements in a set $S$. Intuitively, $\kappa(x)$ is the percentage of mutations in $X_m(x)$ that have a different label $f(x_m)$ from $f(x)$. To abuse the notations, we use $\kappa_{nor}$ and $\kappa_{adv}$ to denote the average sensitivity of normal samples and that of adversarial samples respectively. Our observation is that
\centerline{\framebox{$\kappa_{adv}$ is significantly larger than $\kappa_{nor}$.}} Table~\ref{tb:ratio} shows the $\kappa_{nor}$ and $\kappa_{adv}$ obtained based on the MNIST 
and CIFAR10 dataset with different kinds of attacks in Clverhans~\cite{papernot2017cleverhans} including FGSM, C\&W, JSMA and Black-box. Note that in the case of the last column \emph{wrongly-labeled}, we use those samples in the dataset which are mis-labeled by the DNN $f$ as ``adversarial samples''. Column \emph{StepSize} is a measure of the amount of perturbations applied to the provided sample. We can observe that $\kappa_{nor}$ remains a small value which is comparable to the train error, whereas $\kappa_{adv}$ is much larger than $\kappa_{nor}$ for all adversarial samples generated through different methods.

\begin{table}[t]
\centering
\caption{The confidence interval (99\% significance level) of $\kappa_{nor}$ and $\kappa_{adv}$ of 500 images randomly drawn from MNIST and CIFAR10 dataset with 1000 mutations each under different attacks.}
\begin{adjustbox}{width=\textwidth}
\begin{tabular}{c|c|c|ccccc}
 \toprule
 \multirow{2}{*}{Dataset}  & \multirow{2}{*}{$StepSize$} & \multirow{2}{*}{$\kappa_{nor}$} & \multicolumn{5}{c}{$\kappa_{adv}$} \\ \cline{4-8}
 & & & FGSM & C\&W & JSMA & Black-box & wrongly-labeled\\
 \midrule
 \multirow{3}{*}{MNIST} & 1 & $\SI{0.13 \pm 0.04}{\percent}$ & $\SI{4.86 \pm 0.61}{\percent}$ & $\SI{10.41 \pm 1.39}{\percent}$ & $\SI{6.85 \pm 0.58}{\percent}$ & $\SI{5.98 \pm 2.78}{\percent}$ & $\SI{3.68 \pm 0.65}{\percent}$\\
  & 5 & $\SI{3 \pm 0.16}{\percent}$ & $\SI{14.82 \pm 0.86}{\percent}$ & $\SI{20.6 \pm 1.41}{\percent}$ &$\SI{15.91 \pm 0.93}{\percent}$ & $\SI{13.2 \pm 2.4}{\percent}$ & $\SI{10.32 \pm 0.98}{\percent}$\\
  & 10 & $\SI{6.31 \pm 0.21}{\percent}$ & $\SI{19.34 \pm 1.07}{\percent}$ & $\SI{27.07 \pm 1.51}{\percent}$ & $\SI{21.55 \pm 1.11}{\percent}$ & $\SI{17.05 \pm 2.36}{\percent}$ & $\SI{14.47 \pm 1.15}{\percent}$\\
  \midrule
 \multirow{3}{*}{CIFAR10} & 1 &  $\SI{20.91 \pm 3.27}{\percent}$         & $\SI{61.95 \pm 5.87}{\percent}$ & $\SI{54.06 \pm 3.45}{\percent}$   & $\SI{69.27 \pm 2.64}{\percent}$ & $\SI{62 \pm 7.03}{\percent}$ & $\SI{42.27 \pm 3.12}{\percent}$      \\
  & 5 & $\SI{24.3 \pm 3.04}{\percent}$ & $\SI{63.38 \pm 5.24}{\percent}$ & $\SI{56.91 \pm 3.13}{\percent}$ & $\SI{67.84 \pm 2.86}{\percent}$ & $\SI{63.55 \pm 6.29}{\percent}$ & $\SI{47.6 \pm 2.94}{\percent}$ \\
  & 10 & $\SI{28.3 \pm 2.87}{\percent}$ & $\SI{64.5 \pm 4.86}{\percent}$ & $\SI{59.3 \pm 2.87}{\percent}$ & $\SI{66.15 \pm 3.16}{\percent}$ & $\SI{64.92 \pm 5.72}{\percent}$ & $\SI{48.23 \pm 2.56}{\percent}$\\
 \bottomrule
 \end{tabular}
\end{adjustbox}
 \label{tb:ratio}
\end{table}

\paragraph{Explanatory model}
Next, we aim to provide some intuition on the difference between $\kappa_{nor}$ and $\kappa_{adv}$. 
Given a normal sample $x$, let $R(x, r)$ be a region around $x$ which contains all the possible mutations of $x$ within a certain distance $r$. 
Our hypothesis is that the density of adversarial samples in the region is significantly lower than the density of the normal samples. Thus, if we randomly draw samples around an adversarial sample, we have much higher probability of sampling a normal sample, which explains why $\kappa_{adv}$ is often much larger than the training error. It is easy to see that if the region contains all possible samples, then $\kappa_{nor}$ should be close to the training error.

\begin{figure}[t]
\centering
\includegraphics[width=.65\textwidth]{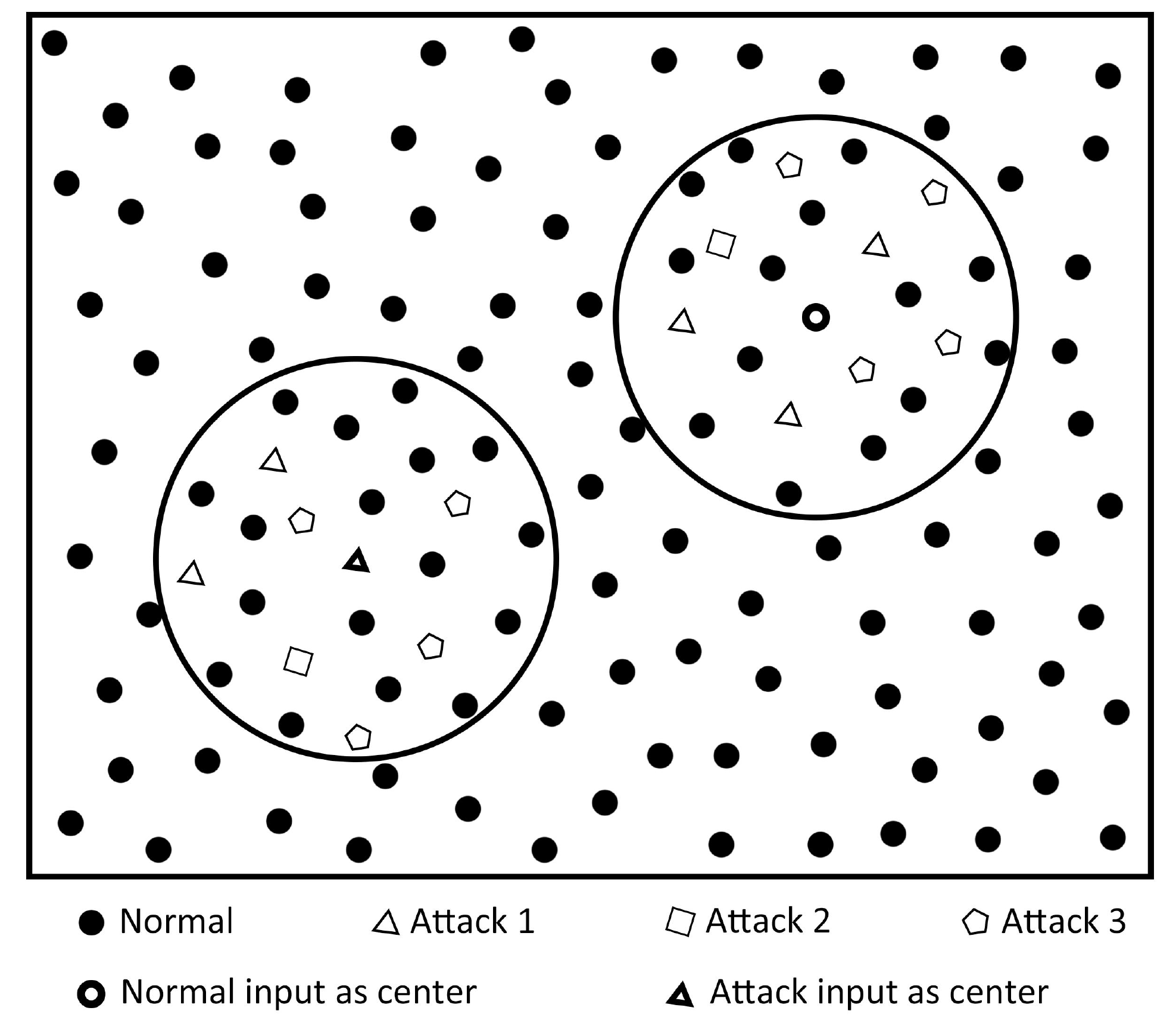}
\caption{Our explanatory model of the mutation testing effect.}
\label{fig:model}
\end{figure}

This simple explanation is illustrated in Figure~\ref{fig:model}, which shows the distribution of normal samples and three classes of adversarial samples crafted from different attacking methods in the input space. The two circles are the mutation region centered around a normal sample and an adversarial sample respectively. The densities of adversarial samples in both circles are expected to be much lower than that of normal samples for a well-trained DNN. Thus, it is much easier to sample a normal sample around an adversarial sample than sampling an adversarial sample around a normal one, which could be an explanation of our observation. Note that adversarial samples outside the two circles are omitted for simplicity.

\section{Adversary detection}\label{defense}

\begin{figure}[t]
\centering
\includegraphics[width=.65\textwidth]{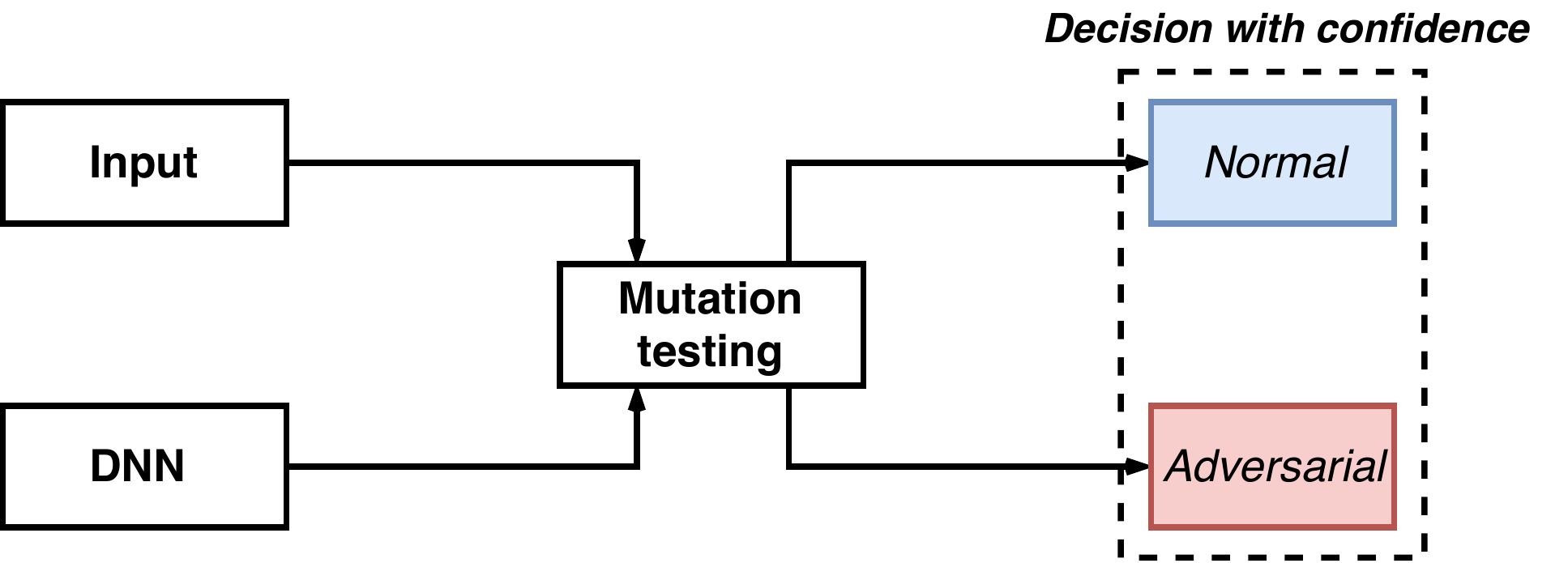}
\caption{Our adversarial input detection framework.}
\label{fig:frame}
\end{figure}

We design our adversary sample detection algorithm based on the above-mentioned observation. The overview of our approach is shown in Figure~\ref{fig:frame}. Given an input $x$ to a DNN, we determine whether it is a normal sample (i.e., $f(x)=c_x$) or an adversarial one (i.e., $f(x)\neq c_x$) through mutation testing. Our detection algorithm further reports a confidence on the detection result. Once $x$ is determined to be an adversarial sample, we raise an alarm and avoid making wrong decisions that could lead to severe consequences. In the following, we introduce how our algorithm works.

\begin{algorithm}[t]
\caption{$DetectAdv(x,f,\kappa_1,\mu,\alpha,\beta,\sigma, StepSize)$}
\label{alg:main}
Let $stop = false$\;
Let $c=0$ be the count of mutations that satisfy $f(x_m)\neq f(x)$\;
Let $n=0$ be the count of total mutations generated so far\;
\While {!stop} {
  Randomly generate a mutation $x_m$ of $x$ with step size $StepSize$\;
  $n=n+1$\;
  \If {$f(x_m)\neq f(x)$}{
    $c=c+1$\;
    Calculate the SPRT probability ratio as $pr$\;
    \If {$pr\ge \frac{1-\beta}{\alpha}$}{
      Accept the hypothesis that $\kappa(x)>\mu\cdot\kappa_1$ and report the input as an adversarial input with error bounded by $\beta$\;
      \Return \;
    }
    \If {$pr\le \frac{\beta}{1-\alpha}$} {
      Accept the hypothesis that $\kappa(x)\le\mu\cdot\kappa_1$ and report the input as a normal input with error bounded by $\alpha$\;
      \Return \;
    }
  }
}
\end{algorithm}

\textbf{The algorithm} 
Our algorithm takes a parameter $\kappa_1$ which is a threshold of sensitivity for normal samples, a $StepSize$ which is a measure of the amount of perturbations, along with multiple parameters required by hypothesis testing.
The basic idea of our algorithm is to use acceptance sampling to test the hypothesis that $\kappa(x)>\mu\cdot\kappa_1$ with strength $\alpha$, $\beta$ and $\sigma$, where $\mu$ is a hyper parameter, $\alpha$, $\beta$, and $\sigma$ are the parameters controlling the strength and indifference region of the test. Recall that $\kappa(x)$ is the measured sensitivity of sample $x$.  The detailed algorithm is shown in Algorithm~\ref{alg:main}.

There are two possible outcomes. If the hypothesis is accepted, it means that the sample has a higher sensitivity than a normal sample. Thus, we report that the input $x$ is an adversarial sample with error bounded by $\beta$. If the hypothesis is rejected, we report that the input is a normal sample with error bounded by $\alpha$. If the test does not satisfy the stopping criteria, the algorithm continues to randomly generate a mutation of the provided sample at line 5. The stopping criteria is calculated at line 9 whenever we observe a label change. In this algorithm, we apply the Sequential Probability Ratio Test (SPRT) to control the sampling procedure~\cite{sprt}, where $pr$ is calculated as
\[pr=\frac{p_1^{c}(1-p_1)^{n-c}}{p_0^{c}(1-p_0)^{n-c}}\] , with $p_1=\mu\cdot\kappa_1+\sigma$ and $p_0=\mu\cdot\kappa_1-\sigma$. The algorithm stops whenever a hypothesis is accepted either at line 11 or line 14. We remark that SPRT is guaranteed to terminate with probability 1~\cite{sprt}. On termination of Algorithm~\ref{alg:main}, we either report the sample as a normal one or an adversarial one with an error bound.

\textbf{Parameter selection} Next, we discuss how to set the value of $\kappa_1$. First of all, we choose to test against $\kappa_{nor}$ instead of $\kappa_{adv}$ because $\kappa_{nor}$ is relatively stable for a given DNN, which is related to the training error and can be empirically estimated using the training data. In contrast, $\kappa_{adv}$ may vary from attack to attack. The hyper parameter $\mu$ reflects the distinction between $\kappa_{adv}$ and $\kappa_{nor}$. To detect a strong attack with $\kappa_{adv}$ close to $\kappa_{nor}$, we would need a small $\mu$, and vice versa. In other words, our algorithm provides a potential measure of the strength of different attacks by measuring the distance between $\kappa_{nor}$ and $\kappa_{adv}$ for a specific attack. If the distance is large, our algorithm is able to quickly detect an adversarial sample by setting a large $\mu$. If the distance is small, our algorithm may have to use a small $\mu$ and evaluate more mutations to reach a conclusion. The parameters for SPRT are decided by how much testing resource we have, a small error bound and indifference region would require more mutations in general. Lastly, the parameter $StepSize$ controls how large is the region that we generate mutations from and thus leads to different $\kappa_{nor}$ and $\kappa_{adv}$. A step size is considered to be optimal if it induces the largest distance between the resultant $\kappa_{nor}$ and $\kappa_{adv}$, which allows the algorithm to terminate early.

\section{Experiments}\label{exp}

There are several goals we want to achieve through the experiments. First of all, we aim to evaluate our hypothesis, i.e., there is a significant difference between $\kappa_{nor}$ and $\kappa_{adv}$ (for different attacks). Secondly, we aim to show that our detection algorithm is able to detect adversarial samples effectively and efficiently. We also aim to provide some practical guidance on the choice of parameters. Lastly, we aim to show that our algorithm improves sample labeling in general. We adopt the MNIST and CIFAR10 datasets, and the set of attacks in Cleverhans~\cite{papernot2017cleverhans} for our experimental evaluation.

\paragraph{Hypothesis evaluation} Table~\ref{tb:ratio} shows the $\kappa_{nor}$ values for normal samples and $\kappa_{adv}$ values for adversarial samples obtained from multiple recently proposed attacking methods in Cleverhans~\cite{papernot2017cleverhans}, which includes the fast gradient sign method (FGSM)~\cite{goodfellow2014explaining}, the C\&W attack (C\&W)~\cite{carlini2017towards}, the Jacobian-based saliency map approach (JSMA)~\cite{papernot2016limitations}, and a practical black-box attack (BB)~\cite{papernot2017practical}. For each dataset, we first randomly draw 500 images from normal samples and attempt to craft 500 adversarial samples using each of the above attacks (notice that not all attempts are successfully and thus the actual number of adversarial samples is less or equal to 500). Then for each image $x$ (either normal or adversarial), we randomly generate 1000 mutations to calculate its sensitivity $\kappa(x)$. We obtain the confidence interval of $\kappa_{nor}$ and $\kappa_{adv}$ by averaging $\kappa(x)$ over all the images. We have the following observations which support our hypothesis and the proposed explanatory model in Section~\ref{mt}.

\emph{Firstly, $\kappa_{nor}$ is a small value comparable to the training error for a well trained DNN, whereas $\kappa_{adv}$ is significantly larger than $\kappa_{nor}$ for all the experimented attacks.} This is especially true when we set a small $StepSize$ (like 1) for generating mutations, in which case the four evaluated attacks has a $\kappa_{adv}$ which is 38, 72, 53 and 46 times larger than $\kappa_{nor}$ respectively for MNIST dataset.

\begin{figure}[t]
    \centering
    \begin{subfigure}[b]{0.49\textwidth}
        \includegraphics[width=\textwidth]{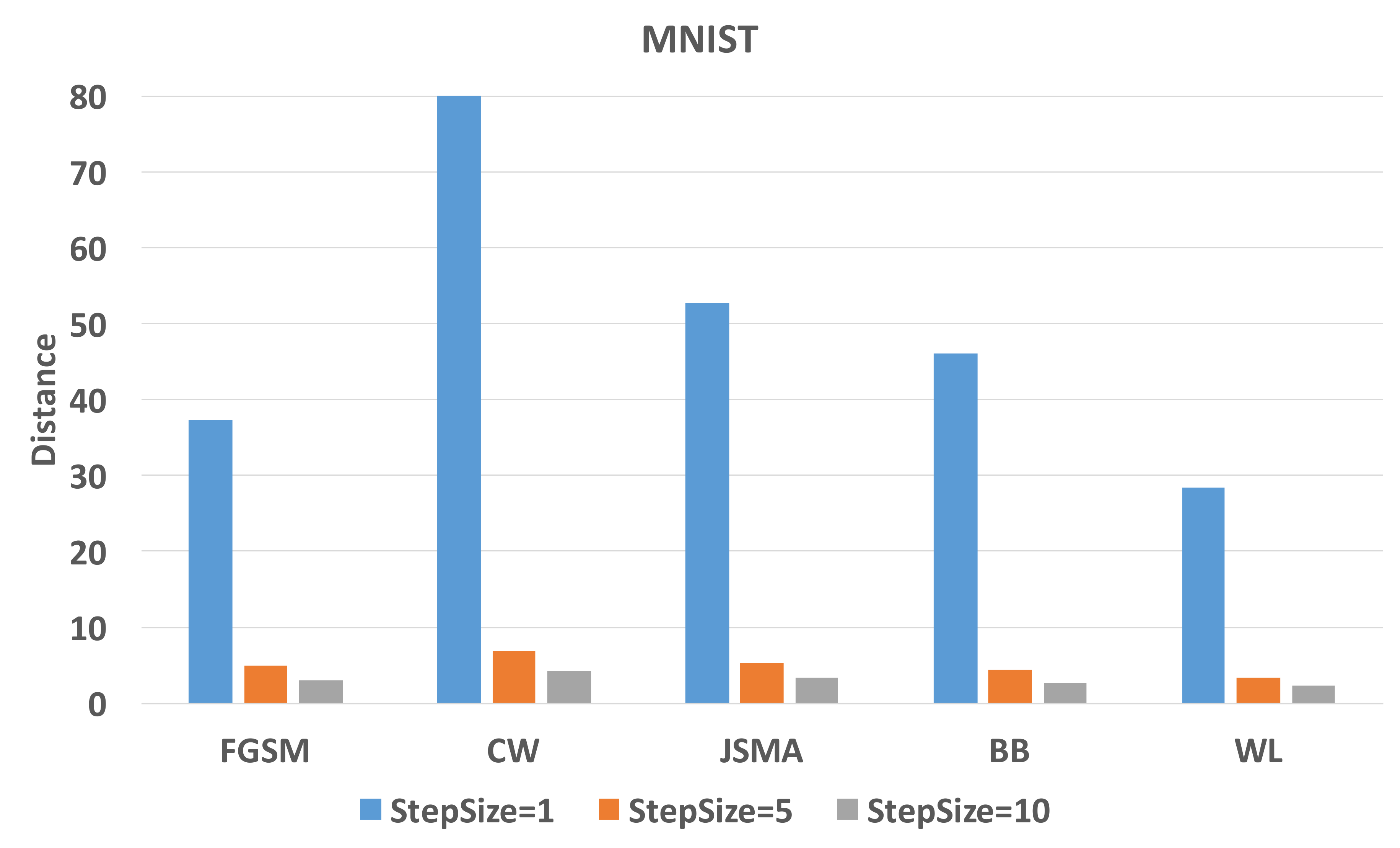}
        \label{fig:dis:mnist}
    \end{subfigure}
    ~ 
    \begin{subfigure}[b]{0.49\textwidth}
        \includegraphics[width=\textwidth]{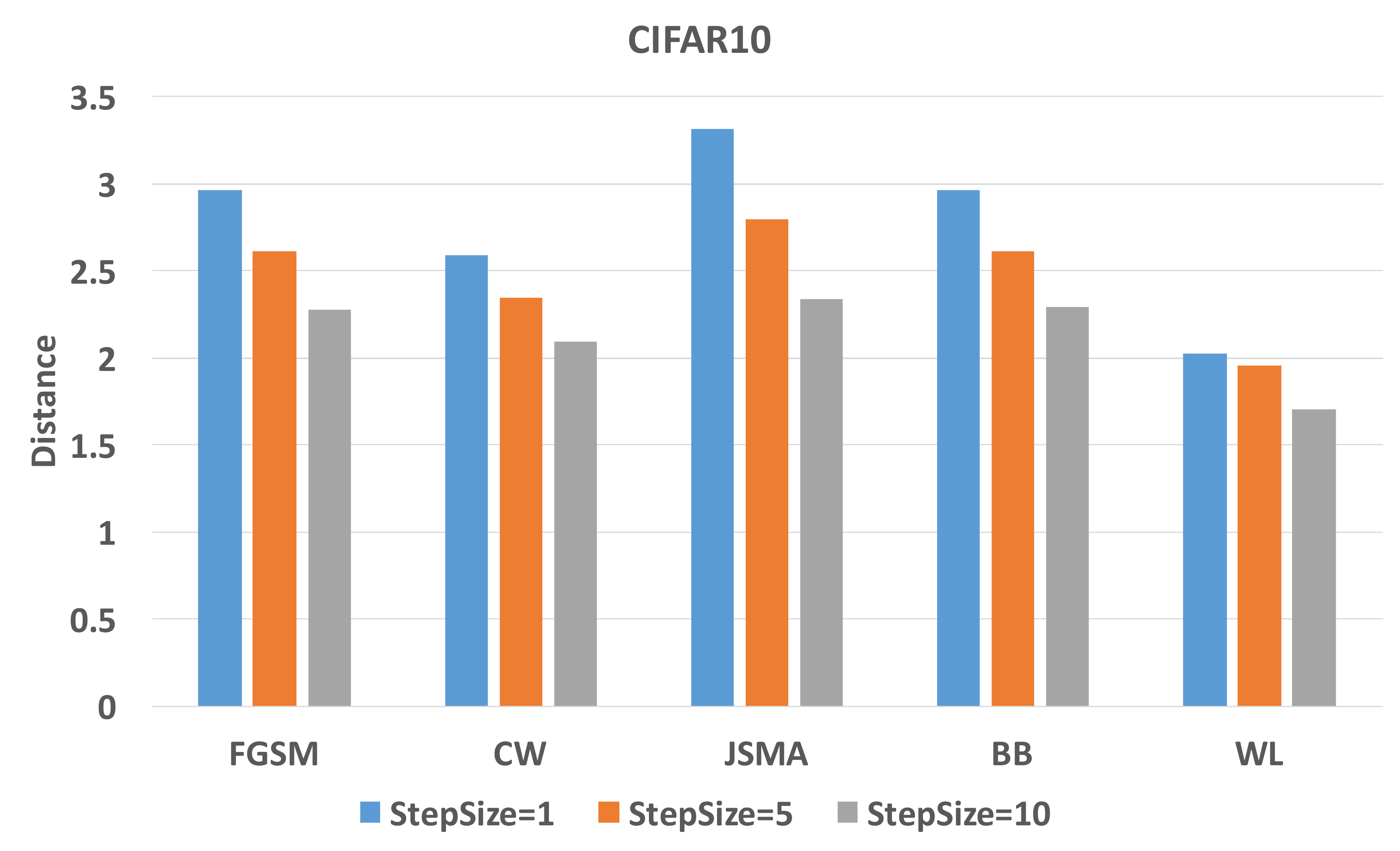}
        \label{fig:dis:cifar10}
    \end{subfigure}
  \caption{The distance between $\kappa_{nor}$ and $\kappa_{adv}$ with different $StepSize$.}
  \label{fig:dis}
\end{figure}

\emph{Secondly, as we increase the $StepSize$ for generating mutations, both $\kappa_{nor}$ and $\kappa_{adv}$ increase (expectedly) and the relative distance between $\kappa_{nor}$ and $\kappa_{adv}$ reduces.} In the following, we measure the relative distance between $\kappa_{nor}$ and $\kappa_{adv}$ using their ratio $\frac{\kappa_{adv}}{\kappa_{nor}}$. Figure~\ref{fig:dis} shows the change of the distance between $\kappa_{adv}$ and $\kappa_{nor}$ as we change the $StepSize$ from 1,5 and 10 for mutation. We can observe that for all the experimented attacks, a smaller $StepSize$ results in a larger distance between $\kappa_{adv}$ and $\kappa_{nor}$ (especially for MNIST). This suggests that we should prefer to use a small $StepSize$ in our algorithm to distinct $\kappa_{nor}$ and $\kappa_{adv}$ so that our algorithm can potentially terminate early.


\textit{Thirdly, adversarial samples crafted by different attacks have different sensitivity to random perturbations.} We can observe that different attack methods have different $\kappa_{adv}$ values (although all of them are significantly larger than $\kappa_{nor}$). As a result, if there is a new attack method, unlikely we are able to know its $\kappa_{adv}$ value. Thus, it is not a good idea to test against $\kappa_{adv}$. However, since 1) $\kappa_{nor}$ is a relatively stable value for a given DNN, and 2) we know that $\kappa_{adv}$ is significantly larger than $\kappa_{nor}$, we may still be able to detect a new attack by testing against $\kappa_{nor}$.

\paragraph{Adversarial sample detection} Given an input $x$ to a DNN, we evaluate our adversarial sample detection algorithm using the following metrics. Firstly, how effective is the detection algorithm, i.e., what is the accuracy (or percentage) that the algorithm successfully identifies an adversarial or a normal sample? Secondly, how efficient is the detection algorithm, i.e., how many mutations do we need in order to reach the conclusion?

\begin{table}[t]
\centering
\caption{Detection results of 4 kinds of crafted adversarial samples for 500 attempts, 500 wrongly-labeled samples and 500 normal samples with maximum 2000 mutations for each sample.}
\label{tb:det}
\begin{adjustbox}{max width=.9\textwidth}
\begin{tabular}{@{}c|ccccccccc@{}}
\toprule
Dataset    & Attack    & $\kappa_1$ & $\mu$ & \#Adversary & \#Identified & Accuracy & \#Mutations   &  \#$lc$     \\ \midrule
\multirow{20}{*}{MNIST}      & JSMA  & 0.0017     & 1.2   & 500 & 475    & 95\%    & 13 & 3.6 \\
       & JSMA  & 0.0017     & 1.5 & 500 & 461        & 92.2\%  & 9  & 2.8 \\
       & JSMA  & 0.0017     & 1.8 & 500 & 463        & 92.6\%  & 9  & 2.8 \\
       & JSMA  & 0.0017     & 2    & 500 & 455       & 91\%  & 6  & 1.92 \\
           & C\&W    & 0.0017     & 1.2  & 450 & 436      & 96.9\%        & 13 & 3.7 \\
           & C\&W    & 0.0017     & 1.5   & 450 & 436      & 96.9\%        & 9 & 2.8 \\
           & C\&W    & 0.0017     & 1.8   & 450 & 425      & 94.4\%  & 8 & 2.8 \\
           & C\&W    & 0.0017     & 2  & 450 & 426      & 94.7\%  & 5 & 1.9 \\
           & FGSM & 0.0017     & 1.2 & 452 & 366        & 81\%     & 56 & 1.5 \\
           & FGSM & 0.0017     & 1.5 & 452 & 325        & 71.9\%     & 38 & 1.3 \\
           & FGSM & 0.0017     & 1.8 & 452 & 284        & 62.8\%     & 29 & 1.2 \\
           & FGSM & 0.0017     & 2 &  452 & 282        & 62.4\%     & 27 & 1.2  \\
           & BB    & 0.0017 & 1.2  &  174   &  137 &  78.7\% &  31   &  2.9     \\
           & BB    & 0.0017 & 1.5  &  174   &  125 &  71.8\% &  18   &  2.3     \\
           & BB    & 0.0017 & 1.8  &  174   &  127 &  73\% &  14   &  2.3     \\
           & BB    & 0.0017 & 2  &  174   &  112 &  64.4\% &  10   &  1.8 \\
           & WL    & 0.0017 & 1.2  &  247   &  171  &  69.2\% &  60   &  1.5     \\
           & WL    & 0.0017 & 1.5  &  247   &  149  &  60.3\% &  38   &  1.3     \\
           & WL    & 0.0017 & 1.8  &  247   &  146  &  59.1\% &  31   &  1.1     \\
           & WL    & 0.0017 & 2 &  247   &  134  & 54.2\% &  28 &  1.2 \\
           & normal    & 0.0017 & 1.2 &  0   &  18 & 96.4\% &  132   &  0 \\
           & normal    & 0.0017 & 1.5 &  0   &  16 & 96.8\% &  84   &  0 \\
           & normal    & 0.0017 & 1.8 &  0   &  14 & 97.2\% &  66   &  0 \\
           & normal    & 0.0017 & 2   &  0   &  10 & 98\% &  58   &  0 \\
           \midrule
\multirow{20}{*}{CIFAR10}    & JSMA  & 0.2418     & 1.01    & 299 & 230    & 76.9\%  & 72  & 25.6 \\
             & JSMA  & 0.2418     & 1.02   & 299 & 226      & 75.6\%    & 52 & 18.5 \\
             & JSMA  & 0.2418     & 1.03  & 299 & 223       & 74.6\%    & 42 & 15.3 \\
             & JSMA  & 0.2418     & 1.04     & 299 & 223    & 74.6\%  & 37  & 13.5 \\
             
           & C\&W    & 0.2418     & 1.01  & 379 & 215        & 56.7\%        & 80 & 27.2 \\
           & C\&W    & 0.2418     & 1.02  & 379 & 213        & 56.2\%        & 56 & 19.3 \\
           & C\&W    & 0.2418     & 1.03  & 379 & 210        & 55.4\%        & 43 & 15.4 \\
           & C\&W    & 0.2418     & 1.04  & 379 & 214        & 56.4\%        & 36 & 13.2 \\
           & FGSM & 0.2418     & 1.01  & 129 & 92        & 71.3\%     & 68 & 24.6 \\
           & FGSM & 0.2418     & 1.02  & 129 & 94        & 72.9\%     & 54 & 19.2 \\
           & FGSM & 0.2418     & 1.03  & 129 & 90        & 69.8\%     & 41 & 15 \\
           & FGSM & 0.2418     & 1.04  & 129 & 91        & 70.5\%     & 35 & 13.1 \\
           & BB    & 0.2418 & 1.01 & 99   &  73 &  73.7\% &  69   &  24.7     \\
           & BB    & 0.2418 & 1.02 & 99   &  71 &  71.7\% &  51   &  18.8     \\
           & BB    & 0.2418 & 1.03 & 99   &  70 &  70.7\% &  41   &  15     \\
           & BB    & 0.2418 & 1.04 & 99   &  67 &  67.7\% &  35   &  13.1 \\
           & WL    & 0.2418 & 1.01  &  500   &  445 &  88.8\% &  61   &  23.6     \\
           & WL    & 0.2418 & 1.02  &  500   &  442 &  88.2\% &  43   &  16.9     \\
           & WL    & 0.2418 & 1.03  &  500   &  437 &  87.2\% &  35   &  14     \\
           & WL    & 0.2418 & 1.04  &  500   &  441 &  88\% &  32   &  12.5 \\
           & normal    & 0.2418 & 1.01 &  0   &  10 & 98\% &  48   &  2.3 \\
           & normal    & 0.2418 & 1.02 &  0   &  11 & 97.8\% &  34   &  1.7 \\
           & normal    & 0.2418 & 1.03 &  0   &  11 & 97.8\% &  27   &  1.2 \\
           & normal    & 0.2418 & 1.04 &  0   &  9  & 98.2\% &  24   &  1.1 \\
           \bottomrule
\end{tabular}
\end{adjustbox}
\end{table}

We run our adversary detection algorithm against a set of adversarial samples crafted using different attacks. We set the parameters in Algorithm~\ref{alg:main} as follows. We set $StepSize$ to be 1 since it induces largest distance between $\kappa_{nor}$ and $\kappa_{adv}$. We set $\kappa_1$ to be the upper bound of the obtained confidence interval of $\kappa_{nor}$: 0.0017 for MNIST ($StepSize=1$) and 0.2418 for CIFAR10 ($StepSize=1$). We vary $\mu$ to see its effect. We set the parameters of hypothesis testing to be $\alpha=0.05$, $\beta=0.05$ and $\sigma=(\mu-1)\cdot\kappa_1$, which gives us a 95\% confidence when a decision is made. Notice that we can use a smaller $\alpha$ and $\beta$ to obtain a higher confidence.

Table~\ref{tb:det} shows the detailed results of our detection experiment. Column `$\kappa_1$' and `$\mu$' are the algorithm parameters. Column `\#Adversary' is the number of successfully crafted adversarial samples from 500 attempts. Column `\#Identified' is the number of detected adversarial samples by our algorithm. Column `Accuracy' shows the accuracy of our detection. Column `\#Mutations' shows the average number of mutations needed for a successful detection. The last column shows the average number of label change during the detection algorithm.
On average, our algorithm achieves an accuracy of 77.4\% (for MNIST) and 76.6\% (for CIFAR10) in terms of detecting the adversarial/normal samples with an average 33 mutations (for MNIST) and 46.5 mutations (for CIFAR10) for each sample, which is both effective and efficient as randomly mutating and obtaining the label of a sample is cheap. 

\emph{Effect of $\mu$.} Intuitively, $\mu$ reflects how confident we are that $\kappa_{adv}$ is significantly larger than $\kappa_{nor}$. A smaller $\mu$ is preferred if we are more conservative. As we increase $\mu$, we are able to detect adversaries with fewer number of mutations. However, the accuracy of the detection might drop a little bit. 

\emph{Wrongly-labeled samples detection.} Recall that in our definition, there is no difference between adversarial samples and benign samples which are wrongly labeled by the DNN. Thus, our algorithm can be potentially used to detect samples which are wrongly labeled. To verify the hypothesis, we random take 500 images from the training dataset which are wrongly labeled by the trained DNN and evaluate their sensibility to random perturbations as well. The last column of Table~\ref{tb:ratio} shows the average $\kappa(x)$ value of the randomly selected wrongly-labeled samples. We can observe that the sensibility of these wrongly-labeled samples to random perturbations are comparable to those generated by four kinds of attacks, which are significantly larger than the sensibility of normal samples. We also run our algorithm against these wrongly-labeled samples (WL) and report our detection result in Table~\ref{tb:det}. We can observe that our algorithm is able to detect these wrongly-labeled samples effectively and efficiently similar to detecting the adversarial samples. This suggests that wrongly-labeled samples are the same as the adversarial samples in our explanatory model from a statistical point of view.


\emph{Normal samples detection.} We also evaluate our algorithm against normal samples which are correctly labeled by the DNN and the `normal' rows in Table~\ref{tb:det} show the detailed results. We can observe that our algorithm also successfully identifies normal samples with high accuracy, i.e., 97.1\% for MNIST and 97.95\% for CIFAR10 on average.

\paragraph{Discussions} 1) We set the ratio of indifference region and $\kappa_1$ to be $\mu-1$. In this case, the left boundary of the indifference region will be $\kappa_1$, which is the upper bound of the obtained confidence interval of $\kappa_{nor}$. In practice, we could set $\mu$ according to $\kappa_1$ and the size of indifference region. 2) Decreasing $\alpha$ and $\beta$ will improve the accuracy of the detection but require more mutations in general. 3) We achieve different accuracy on different attacks, which could be a potential measure of the strength of the attacks and provide insight of defense against specific attacks. 4) We restrict our mutation generation in a realistic way according to the methods in~\cite{pei2017deepxplore}.

\section{Conclusion and Outlook}\label{con}

In this work, we report our discovery that commonly crafted adversarial samples and normal samples of DNN have significant different sensibility to random perturbations. We then design an algorithm for adversarial sample detection based on mutation testing of an input sample. Our experiments show that our algorithm is able to detect adversarial samples with high accuracy (correct detection) and low cost (few mutations needed). Our work is not limited to a specific kind of attack or a set of available adversarial data, but has the potential to be applied to a wide range of attacks. It can also be potentially applied to re-label those benign samples which are wrongly-labeled by the DNN.

\bibliographystyle{plain}
\bibliography{main.bib}

\end{document}